\documentclass[letterpaper,10pt,conference]{ieeeconf}  %

\makeatletter
    \let\NAT@parse\undefined
\makeatother
\usepackage[square,numbers,sort&compress]{natbib}
\usepackage[table, dvipsnames]{xcolor}
\usepackage[font=small, labelfont=bf]{caption}
\usepackage[colorlinks = true,
            linkcolor = black,
            urlcolor  = blue,
            citecolor = magenta,
            bookmarks = true, pagebackref]{hyperref}
\usepackage{lipsum}
\usepackage{xspace}
\usepackage{todonotes}
\usepackage{amsmath}
\usepackage{amssymb}
\usepackage{booktabs}
\usepackage{array}
\usepackage{soul}
\usepackage{float}
\usepackage{booktabs}
\let\labelindent\relax
\usepackage{enumitem}

\def\eqref#1{equation~\ref{#1}}

\def\1{\bm{1}}

\DeclareMathAlphabet{\mathsfit}{\encodingdefault}{\sfdefault}{m}{sl}
\SetMathAlphabet{\mathsfit}{bold}{\encodingdefault}{\sfdefault}{bx}{n}

\def\gC{{\mathcal{C}}}

\def\gM{{\mathcal{M}}}

\usepackage{balance}

\newcommand{\MyPara}[1]{\noindent\textbf{#1}}
\newcommand{\ours}[0]{\rowcolor{LimeGreen!30}}
\newcommand\blfootnote[1]{%
  \begingroup
  \renewcommand\thefootnote{}\footnote{#1}%
  \addtocounter{footnote}{-1}%
  \endgroup
}

\title{\LARGE\bf GNM: A General Navigation Model to Drive Any Robot}
\newcommand{\MethodName}[0]{GNM\xspace}

\author{Dhruv Shah$^{\dagger\beta}$, Ajay Sridhar$^{\dagger\beta}$, Arjun Bhorkar$^{\beta}$, Noriaki Hirose$^{\beta\tau}$, Sergey Levine$^{\beta}$}

\begin{document}

\makeatletter
\let\@oldmaketitle\@maketitle%
\renewcommand{\@maketitle}{\@oldmaketitle%
    \centering
    \vspace*{1mm}
    \includegraphics[width=\linewidth]{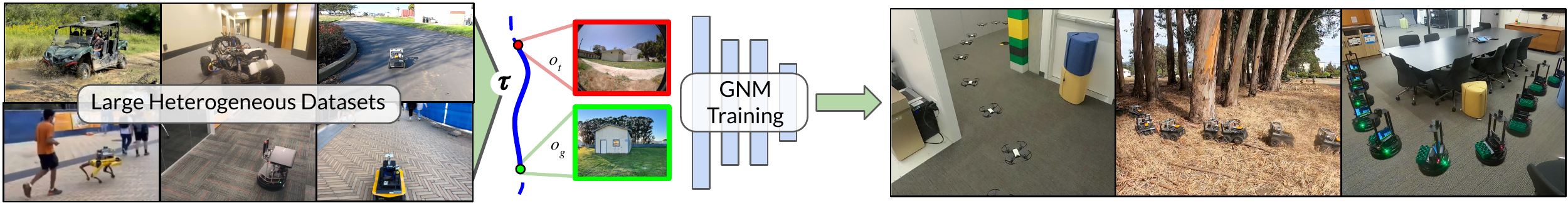}
    \captionof{figure}{\textbf{A general navigation model to drive any robot.} By training on diverse, heterogeneous datasets, a single ``omnipolicy'' can control a variety of robots in challenging environments, including \emph{new} robots, without any robot-specific data collection.}
    \label{fig:teaser}
    \vspace*{-3mm}
}
\makeatother
\maketitle
\IEEEpeerreviewmaketitle
\setcounter{figure}{1}
 
\maketitle

\begin{abstract}
Learning provides a powerful tool for vision-based navigation, but the capabilities of learning-based policies are constrained by limited training data. If we could combine data from all available sources, including multiple kinds of robots, we could train more powerful navigation models. In this paper, we study how a \emph{general} goal-conditioned model for vision-based navigation can be trained on data obtained from many distinct but structurally similar robots, and enable broad generalization across environments and embodiments. We analyze the necessary design decisions for effective data sharing across robots, including the use of temporal context and standardized action spaces, and demonstrate that an \emph{omnipolicy} trained from heterogeneous datasets outperforms policies trained on any single dataset. We curate 60 hours of navigation trajectories from 6 distinct robots, and deploy the trained \MethodName on a range of new robots, including an underactuated quadrotor. We find that training on diverse data leads to robustness against degradation in sensing and actuation.
Using a \emph{pre-trained} navigation model with broad generalization capabilities can bootstrap applications on novel robots going forward, and we hope that the GNM represents a step in that direction.
For more information on the datasets, code, and videos, please check out our project page\footnote{\href{http://sites.google.com/view/drive-any-robot}{\texttt{
sites.google.com/view/drive-any-robot}}}.

\end{abstract}

\blfootnote{$^\dagger$ Equal Contribution. $^{\beta}$UC Berkeley, $^{\tau}$Toyota Motor North America.}

\vspace*{-1em}
\section{Introduction} \label{sec:introduction}

Machine learning methods have enabled broad generalization with real-world applicability in natural language processing~\cite{Radford2018ImprovingLU}, visual perception~\cite{deng2009imagenet, carreira2017kinetics, grauman2022ego4d}, and other domains~\cite{ramesh2022dalle2, chen2021codex} by leveraging Internet-scale data. Such generalization typically requires learning general patterns from diverse datasets, which are usually collected once and then reused for various purposes. Such large-scale models also support the ability to be adapted for new tasks by reusing the representations learned from broader, larger, and more general datasets, for example by or zero-shot transfer~\cite{chen2020simple, shah2022robotic, wei2022finetuned}, or fine-tuning on target-domain data. Although this paradigm has been very successful, it is difficult to apply in robotics due to the sheer diversity of environments and platforms across researchers. Control policies learned end-to-end usually require separate data collection for each robotic platform, leading to ``fragmentation'' in progress, where every researcher works with their own robot-specific dataset and policies, making it infeasible to accumulate large enough datasets. Can we overcome this challenge by training models on more general and reusable cross-robot datasets?

We study this question in the context of visual navigation, where heterogeneity between robots might include different camera hardware, viewpoints, dynamics, and more broadly, embodiments, but where the over-arching navigation objective looks similar irrespective of these differences.
A wheeled robot, quadruped, or a drone all have the same abstract objectives: to explore the environment, plan a path to the goal, and avoid collisions. Leveraging this shared abstraction across robots and training a general navigational \emph{omnipolicy} from large-scale data could enable broad generalization to novel environments, unseen sensor parameters (e.g., camera intrinsics and extrinsics), and new robot configurations. %

In this paper, we propose to take a step towards this kind of data sharing by training an embodiment-agnostic general navigation model (GNM) from an aggregated multi-robot dataset.  The primary contribution of our work is a framework for training a \emph{general} omnipolicy from multi-robot datasets, with empirical evidence that such an omnipolicy can effectively learn from heterogeneous datasets and generalize to novel robot platforms. To facilitate this, we aggregate a large heterogeneous dataset of navigation trajectories collected across 6 robots, spanning 60 hours of interactions in challenging indoor and outdoor environments. We train the GNM on this dataset and deploy it on 4 distinct robot platforms, including 2 new robots.
We show that a single learned policy can be used across multiple robots to perform goal-reaching in challenging indoor and outdoor environments, outperforming policies trained with any single dataset. We also report robustness to degradation in camera parameters, tire damage, and other gradual changes that the robot may experience over its lifetime. 

We have publicly released the trained \MethodName policy, code used to train and deploy our models on various popular robot platforms, as well as the dataset used to train these models at our project page. 
We hope that this represents a step towards both general-purpose multi-robot datasets and general-purpose \emph{visual navigational models} that can be deployed on a wide range of robots --- similar to how practitioners currently use pre-trained models in vision and language, such models could constitute pre-trained backbones for visual navigation. %

\section{Related Work}
\label{sec:relatedwork}

Learning from large, diverse robotic datasets has been studied for various robotic applications where data sharing across \emph{similar} robots helps scale learning to challenging environments~\cite{devin2017multi, dasari2020robonet, Yu2020bdd100k}. However, for applications such as ground or aerial navigation, with different sensors and robot dynamics, current approaches tend to rely on learning from small datasets which are only representative of a single robotic platform. Our paper proposes learning navigation behavior from heterogeneous robot datasets, collected across multiple embodiments.

Our work is closely related to transfer learning, where the objective is to train policies that transfer across domains, such as across dynamics~\cite{yu2017preparing, peng2018simtoreal, feng2022genloco}, environments~\cite{kumar2021rma}, morphologies~\cite{gupta2017learning, huang2020one, kang2021hierarchically}, viewpoints~\cite{sadeghi2018viewpoint}, and embodiments~\cite{hirose2022exaug}. Our focus is not on designing specific domain adaptation algorithms or hand-engineered augmentations~\cite{hirose2022exaug} for transfer, but rather studying how direct generalization of simple, high-capacity models trained on real-world data can provide a path to broadly applicable navigational policies. Towards this, our work is also closely related to DroNet~\cite{loquercio2018dronet}, which imitates expert on-road driving data to control a quadrotor. We take this paradigm one step further, showing that we can train \emph{goal-conditioned} policies on data from \emph{multiple} robots and control new ones, including a quadrotor.%

Prior work has also explored learning of visual representations or end-to-end policies from passive data, such as YouTube videos, which can be scaled up massively without real-world data collection~\cite{chang2020semantic,  hahn2021no, nair2022rm, radosavovic2022realworld}. We explore a complementary direction, studying how readily available on-robot data (also passive) can lead to generalizable policies. This is particularly relevant for navigation, where data is plentiful, and trajectories from multiple robots can directly train a policy, as opposed to two-stage methods that use Internet data for representation learning followed by in-domain adaptation.

Following a large body of research in visual navigation~\cite{hirose2019deep, shah2020ving, chaplot2020nts, meng2020scaling, savinov2018sptm}, we use a combination of topological graphs for high-level planning and image-goal policies for low-level control, which gives us an efficient way to scale reactive policies for long-range navigation~\cite{meng2020scaling, eysenbach2019sorb}. 
Prior work has also extended this framework for complex tasks beyond goal-reaching, such as exploration~\cite{chaplot2020nts, shah2021rapid, shah2022viking}, instruction following~\cite{shah2022robotic}, and reinforcement learning~\cite{shah2022offline}. %
We show that that our \MethodName can be coupled with such topological graphs to scale image-goal navigation to new robots.

\begin{table}
\centering
{\footnotesize
\begin{tabular}{cllccl}
\toprule
& Dataset & Platform & Speed & Hrs. & Environment \\ \midrule
1 & GoStanford~\cite{hirose2019deep} & TurtleBot2 & 0.5m/s & 14h & office \\
2 & RECON~\cite{shah2021rapid}  & Jackal & 1m/s & 25h & off-road \\
3 & CoryHall~\cite{kahn2018gcg} & RC Car & 1.2m/s & 2h & hallways \\ %
4 & Berkeley~\cite{shah2022viking}  & Jackal & 2m/s & 4h & suburban \\
5 & SCAND-S~\cite{karnan2022scand} & Spot & 1.5m/s & 8h & sidewalks \\
6 & SCAND-J~\cite{karnan2022scand} & Jackal & 2m/s & 1h & sidewalks \\
7 & Seattle~\cite{shaban2021semantic} & Warthog & 5m/s & 1h & off-road \\
8 & TartanDrive~\cite{triest2022tartan} & ATV & 10m/s & 5h & off-road \\
9 & NeBula~\cite{agha2021nebula} & ATV & 10m/s & 10h & off-road \\
\midrule
& Ours & & & 70h &  \\
\bottomrule
\end{tabular}}
\caption{\textbf{The \MethodName training dataset} contains 70 hours of navigation data in diverse environments across 6 different robots.}
\label{tab:dataset}
\vspace*{-1.8em}
\end{table}

\section{Multi-Robot Training Dataset}
\label{sec:dataset}

Our aim is to train a \emph{general} visual navigation model that can learn broadly applicable navigational affordances across a variety of distinct robotic systems.
To facilitate such large-scale policy learning, we aggregated a heterogeneous dataset of navigation trajectories sourced from 8 datasets collected on %
robotic platforms with varying dynamics, sensors, and behaviors. The datasets contain a variety of challenging indoor and off-road environments (Table~\ref{tab:dataset} and Fig.~\ref{fig:teaser}). We have publicly released this dataset on the project page. %

The \MethodName dataset contains over 60 hours of real-world navigation trajectories: a combination of tele-operated and autonomous navigation behaviors collected across 6 distinct robotic platforms, including 4 commercially available platforms (TurtleBot, Clearpath Jackal, Warthog and Spot) and 2 custom platforms (Yamaha Viking ATV, RC Car). The trajectories contain widely varying robot dynamics and top speeds ranging between 0.2 and 10m/s, operating in a diverse set of environments (e.g., office buildings, hallways, suburban, off-road trails, university campus etc.).

To train navigation policies that can operate solely from egocentric visual observations, the dataset contains forward-facing RGB images paired with the robot's commanded actions and local odometry measurements. Each robot has different camera parameters, necessitating any successful policy to generalize across variations in camera pose and intrinsic parameters, though all platforms use the same type of sensor (monocular RGB camera). It is straightforward to further expand \MethodName by adding other datasets of relevant navigation behaviors~\cite{biswas2013cobot, bianco2016nclt}, %
or mix-and-match subsets of the dataset based on the desired application, %

\section{Training a General Navigation Model}
\label{sec:method_description}

To study a common navigation task across robots and environments, we consider the problem of image-goal navigation~\cite{zhu2016targetdriven}, where a robot is tasked with navigating to a goal location $G$ specified as an image observation $o_G$ taken at $G$. Unlike PointGoal~\cite{anderson2018evaluation}, GPS navigation, or semantic objectives~\cite{wani2020multi}, image-goal navigation is a general framework that does not rely on ground truth localization or semantic labels, and allows us to formulate a very general navigation task that can be trained with any visual navigation dataset. Our goal is to train a goal-reaching policy $\pi(o_t, o_G)$ that can navigate solely from egocentric visual observations. To provide a general task representation for this policy, we condition it on the desired goal $o_G$ and integrate it into a navigational system based on topological graphs~\cite{hirose2019deep, shah2020ving, chaplot2020nts, meng2020scaling}.

Such systems have shown great navigation results in a variety of indoor and outdoor environments --- what would it take to train such a policy \emph{across} robots, with varying controllers, dynamics and sensor placements? We highlight two key ingredients in training multi-robot policies: (i) carefully choosing the right action representation that facilitates transfer across robots,
and (ii) conditioning the policies on a ``summary'' vector that allows it to deduce the properties of the robot it is controlling, so different robots can exhibit different, valid capabilities. Although we found the particular design decisions described in this section to be important for good performance, as we discuss in our experiments (Sec.~\ref{sec:analysis}), we emphasize that the primary contribution of our work is \emph{not} a novel learning algorithm, but an empirical demonstration that policies learned from heterogeneous datasets can generalize broadly to new environments and new robots.

\begin{figure}
    \centering
    \includegraphics[width=\columnwidth]{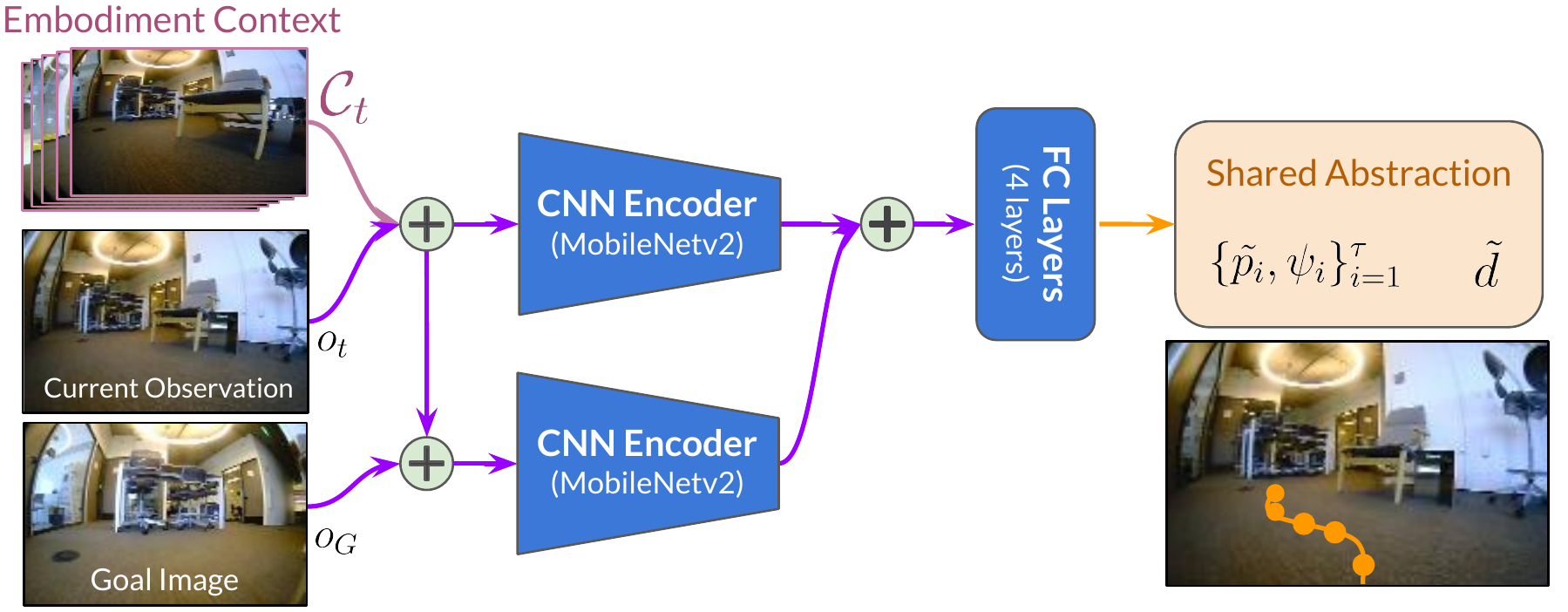}
    \caption{\textbf{\MethodName architecture.} We modify a typical goal-conditioned architecture (purple) by conditioning it on additional context from the target robot (pink) and making predictions in a shared, normalized action space (yellow).}
    \label{fig:model} %
    \vspace*{-1.3em}
\end{figure}

\vspace*{-0.3em}
\subsection{A Shared Abstraction Across Robots}
\label{sec:abstraction}

While the general task of navigation from egocentric images is common across robots, the specific inputs (camera observations) and outputs (actions, dynamics) can vary substantially: a TurtleBot is differential-drive, expects low-level velocity commands, and has a top speed of 0.5m/s, whereas an ATV uses Ackermann steering, expects throttle and steering commands, and drives up to 20$\times$ faster. Learning a common control policy that operates directly on these raw, unstructured outputs can be challenging due to these inconsistencies and high-variance outputs (e.g., speed $\in [0.2, 10]$m/s). This is further exacerbated when generalizing to new robots, where the policy might need to ``guess'' how fast it should move.

To this end, we propose using a shared abstraction to allow the goal-reaching policies to operate in a \emph{transformed} action space that is consistent across robots, making the data points look ``similar'' and easier to learn common patterns from.
In our experiments, we found this to be important to be able to learn from multiple datasets (see Sec.~\ref{sec:analysis_abstraction} for analysis).
We use a combination of \emph{relative} waypoints $p(x,y)$ and yaw change $\psi$ as a mid-level action space.
Labels for these actions can be obtained by using local odometry, which are easily available across datasets. Additionally, the policy also predicts the temporal distance to the goal $d$, as a measure of traversability, which is used by the navigation system to estimate the connectivity of the topological graph.

While this gives a shared action space across robots, we found that the varying dynamics (e.g., different top speeds) across robots can make it challenging for learning algorithms to learn a joint policy. To alleviate this, we propose using a \emph{normalized} action space $\{\tilde{p}(x,y), \psi\}$, where $\tilde{p}:=\frac{1}{\alpha}p$ is scaled by a robot-specific factor $\alpha$ corresponding to the top speed of the robot. The temporal distance $\tilde{d}$ is also estimated in this normalized scale.
Given this abstract action space, a robot-specific controller can be used to (i) \emph{unnormalize} the waypoints, and (ii) \emph{track} them (e.g., PID, MPPI) to extract low-level commands (e.g., velocities or motor commands).

\subsection{Embodiment Context}
\label{sec:context}

When deployed on an arbitrary robot, the policy must infer the capabilities of that particular robot. For instance, a TurtleBot can spin in-place but not go over bumps on the road, whereas an RC Car can easily traverse small bumps but has a limited turning radius. A simple way to provide such awareness to the policy is to condition it on hand-designed parameters that provide a concise ``summary'' of capabilities, such as its size, turning radius etc. Defining these parameters by hand presents a barrier to fast and easy deployment of the policy to new robots, and requires human intuition to identify and define a relevant set of parameters. Instead, we propose a simple and automatic approach: 
rather than manually defining parameters that fully identify the robot, we
use a sequence of consecutive past observations from the robot's viewpoint to infer a learned \emph{embodiment context} $\gC_t$, and condition the learned policy on this context in addition to the observations. This context contains information about the robot's configuration and dynamics, which can be used to condition the behavior of the policy.

While this context may not contain \emph{all} information to fully identify the robot, we hypothesize that it is sufficient to effectively control the robot. Our experiments show that the embodiment context allows the same policy to be deployed on novel robot configurations without designing any hand-engineered robot representation. We empirically evaluate different ways of providing context in Sec.~\ref{sec:analysis_embodiment} and find that the most effective representation is achieved by using a temporally consistent context $\gC_t$ that conditions the policy on $k$ consecutive past observations $\{o_{(t-k):(t-1)}\}$. 

\begin{figure*}[ht]
    \centering
    \includegraphics[width=\linewidth]{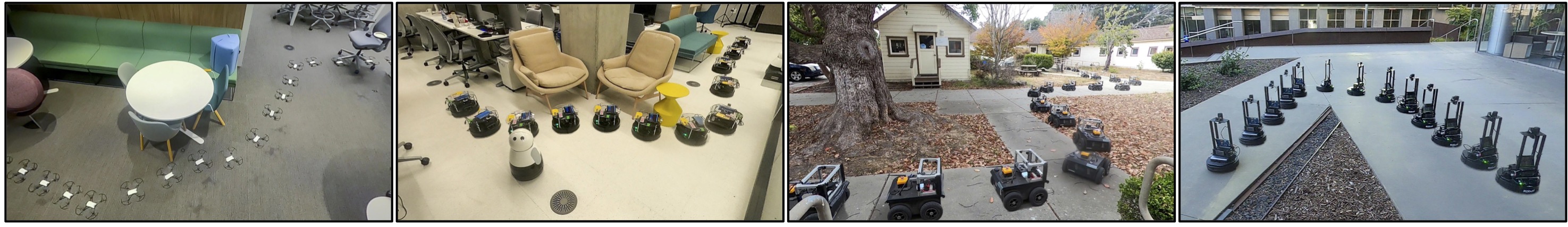}
    \caption{\textbf{Depoying the \MethodName omnipolicy.} We evaluate on 4 different robots in challenging indoor and outdoor environments.}   \label{fig:qualitative_all}
    \vspace*{-1.3em}
\end{figure*}

\subsection{Implementation Details}
\label{sec:implementation}

A combination of conditioning the policies on embodiment context and transforming the action space can allow a simple goal-reaching policy to be trained from heterogeneous datasets. It is important to note that the proposed modifications are orthogonal to the choice of downstream policy architecture and learning algorithm, and we could use different encoders or train with reinforcement learning.

\vspace{1mm}
\MyPara{Architecture:} We use a goal-conditioned policy architecture that takes as input the current observation $o_t$ and goal observation $o_G$, and predicts \emph{normalized} waypoints and distances. Additionally, we condition on temporal context $\gC_t$, which is constructed by stacking the past $k=5$ consecutive observations. Visual inputs to the network are provided as $85\times64$ RGB images for all observations. Following prior work~\cite{shah2021rapid, nair19ccrig}, we train context-conditioned representations by using separate MobileNetv2 encoders for (i) the current observation $\{o_t, \gC_t\}$, and (ii) conditional goal observation, as shown in Fig.~\ref{fig:model}. The two embeddings are concatenated and passed through three fully-connected layers to two prediction heads: normalized temporal distance $\tilde{d}_t$ and a sequence of $\tau=5$ normalized future waypoints $\{ \tilde{p}_i, \psi_i \}_{i=1}^\tau$.

\vspace{1mm}
\MyPara{Training:} Following the procedure of Shah et. al.~\cite{shah2020ving}, we use a combination of image-goal pairs sampled from the same trajectory in the dataset as ``positives'', and ``negatives'' sampled from \emph{different} trajectories,
to obtain training data pairs. The distance head is trained on both positives and negatives, whereas the action head is only trained on positives. We train the two heads jointly with supervised learning using an $\ell_2$ regression loss. We use multi-GPU training with batch sizes between 400--1200 and perform gradient updates using the Adam optimizer~\cite{Kingma2015AdamAM} with a learning rate of $5\times10^{-4}$.

\vspace{1mm}
\MyPara{Deployment:} We combine this goal-reaching policy with a topological map $\gM$, where nodes are represented by the robot's observations (augmented with the embodiment context), and edges are computed using the temporal distance estimates $d$ from the trained policy, following the setup of ViNG~\cite{shah2020ving}.
At every time step, the robot associates
its current and goal observations in $\gM$, i.e., finds the node with smallest temporal distance to it, and computes the optimal sequence of subgoals $\{s_i\}$ using Dijkstra's algorithm. The policy $\pi$ is queried with the current observation $\{o_t, \gC_t\}$ and immediate subgoal $s_1$ to obtain a sequence of waypoints $\{ \tilde{p}_i, \psi_i \}_{i=1}^\tau$, which are tracked by a robot-specific low-level controller.

\section{Deploying the \MethodName Across Robots}
\label{sec:experiments}

We deploy our learned \MethodName omnipolicy in a variety of challenging indoor and outdoor environments on four different robot platforms. We designed out experiments to answer the following questions:
\begin{enumerate}[label={\bf Q\arabic{*}.}, leftmargin=2.8\parindent]
    \item Can multi-robot training enable generalization to \emph{novel} robots and environments?
    \item Do \MethodName policies outperform policies trained solely on single-domain data? 
    \item How important are the design choices made in Sec.~\ref{sec:method_description} for attaining good performance with the \MethodName?
    \item Are policies trained with multiple datasets more robust to degradation than single-domain policies?
\end{enumerate}

\subsection{Meet the Robots}
\label{sec:robots}

We deploy the \MethodName on four distinct robotic platforms, including a quadrotor and two other \emph{novel} robots with no corresponding training data, as shown in Fig~\ref{fig:qualitative_all}.

\vspace{1mm}
\MyPara{Vizbot:} A custom-built robot platform inspired by the design of Niwa et. al.~\cite{niwa2022spatio}, based on a Roomba. It is equipped with an off-the-shelf PCB-mounted fisheye camera. \emph{There is no training data from a Vizbot or any other Roomba-like robot.}

\vspace{1mm}
\MyPara{DJI Tello:} A commercially available quadrotor equipped with a forward-facing camera. \emph{There is no training data from any quadrotor for \MethodName.}
We restrict the drone to a horizontal plane 1m off the ground, to mimic ground navigation.

\vspace{1mm}
\MyPara{Clearpath Jackal UGV:} A commercially available off-road platform equipped with an off-the-shelf PCB-mounted fisheye camera. \emph{This system resembles the  data collection platform used for the RECON, Berkeley, and SCAND-J datasets}, but has a different camera and mounting height.

\vspace{1mm}
\MyPara{LoCoBot:} A popular open-source platform based on a Kobuki, equipped with an off-the-shelf PCB-mounted fisheye camera. \emph{There is no training data from a LoCoBot}, although GS was collected on a similar TurtleBot2, albeit with a different spherical camera at a lower height.

\subsection{Zero-Shot Deployment}
\label{sec:results_main}

Towards answering \textbf{Q1}, we deploy the \emph{same} trained \MethodName on four distinct robotic platforms \emph{without} any fine-tuning per robot. Fig.~\ref{fig:qualitative_all} and Table~\ref{tab:results_summary} summarize our evaluation in a variety of indoor and outdoor environments on 4 different robots, all using the same model. Most notably, the \MethodName can control a Tello, despite never having seen any trajectories from aerial robots in \MethodName. A \MethodName policy consistently outperforms single robot policies across all tested robots, performing up to 5x better in some cases. We also observe generalization to massively out-of-distribution (OOD) settings, like a LoCoBot navigating outdoors on a sidewalk, or a Jackal navigating inside an office building, which were not present in the training data.
This suggests that training on heterogeneous datasets can enable generalization to novel environment-robot pairs, as well as entirely new robots.

To better understand how data sharing benefits performance (\textbf{Q2}), we quantitatively evaluate the navigation performance of policies trained with heterogeneous datasets in an assortment of 20 indoor and outdoor environments on the Jackal and LoCoBot platforms (Tables~\ref{tab:results_locobot}, \ref{tab:results_jackal}). To project the performance trends with varying amounts of data, we train policies from increasingly diverse subsets the training data --- ``Small'', ``Mid'', and ``Large'', corresponding to data from the first 2, 4, and 6 datasets listed in Table~\ref{tab:dataset}. We quantify performance using success rates, measured as the mean progress made towards the goal. For videos of our experiments and more information on the testing environments, please check out the supplementary video and project page.

\begin{figure}
    \centering
    \includegraphics[width=0.95\columnwidth]{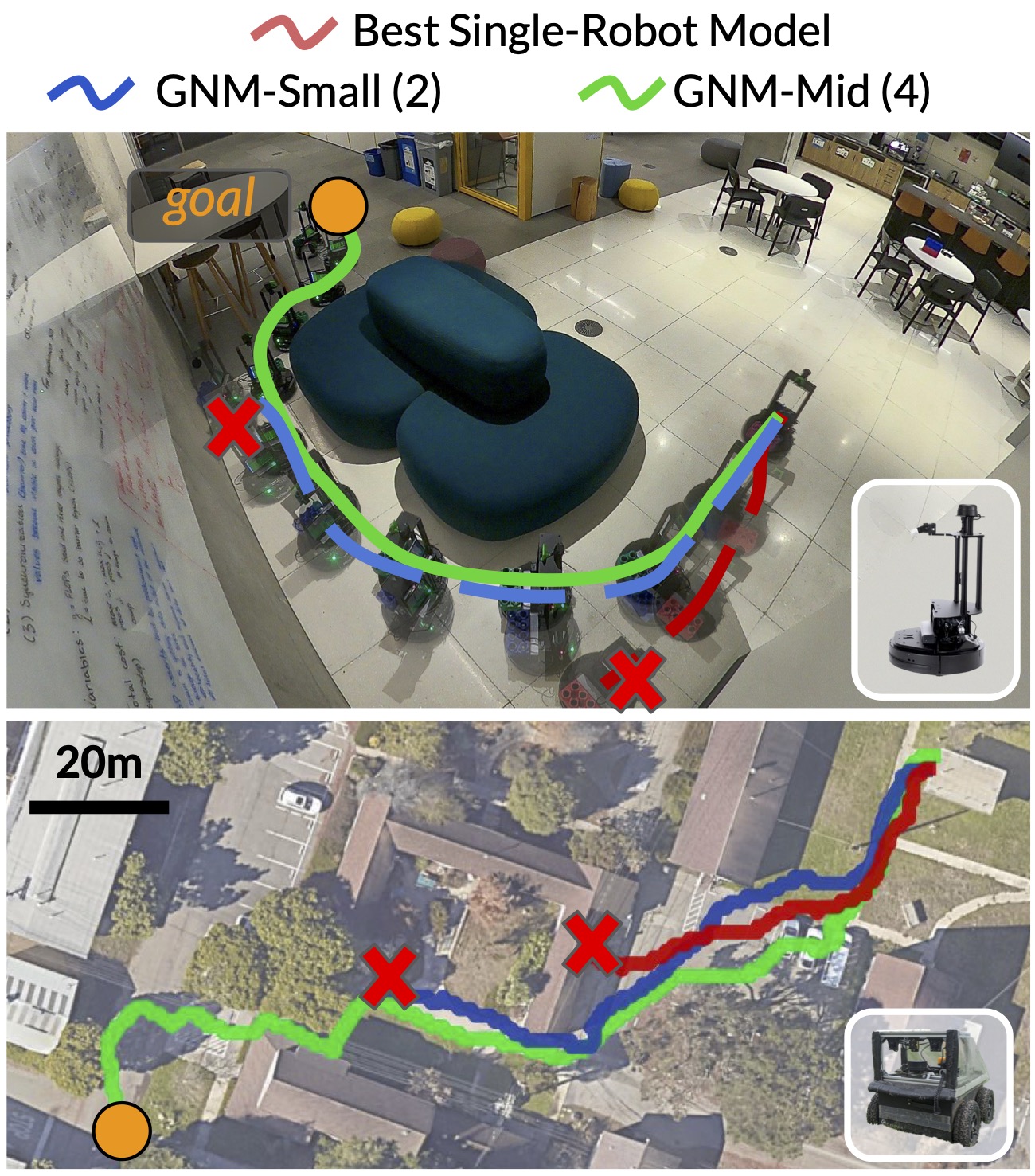}
    \caption{\textbf{Qualitative comparison.} Policies trained with increasingly diverse data lead to better generalization to a LoCoBot (top) and Jackal (bottom). Both robots were controlled by the \emph{same} policy.}
    \label{fig:qualitative_comparison}
\end{figure}

Deploying on a LoCoBot, which is an \emph{unseen robot} with no corresponding data present in the dataset, we find that policies trained on a single dataset (e.g., GoStanford (GS)~\cite{hirose2019deep} or CoryHall~\cite{kahn2018gcg}) fail to generalize to a new embodiment with different sensors. Fine-tuning visual representations trained for task-agnostic datasets like ImageNet, which is a popular strategy for pre-training in many vision-based applications~\cite{yosinski2014transfer, razavian2014cnn}, improves a bit but still struggles in a majority of the environments. However, policies trained by sharing task-relevant datasets across robots significantly outperform these single-domain policies, as shown in Table~\ref{tab:results_locobot}. We also observe that adding more and diverse datasets (\MethodName-Large) contributes towards improvements in performance, despite the additional data coming from seemingly unrelated tasks (e.g., off-road driving). Fig.~\ref{fig:qualitative_comparison} shows an example office environment where increasing the diversity of training data improves performance.

We observe similar trends on a Jackal, which is deployed on a variety of \emph{previously unseen} outdoor and indoor environments (Table~\ref{tab:results_jackal}). Unsurprisingly, a single-domain policy trained on off-road RECON data~\cite{shah2021rapid} performs well for many outdoor environments, but struggles with navigating indoors, which is OOD for the RECON dataset. Similarly, a GS policy struggles in outdoor environments but succeeds in some easy indoor environments. \MethodName omnipolicies are able to generalize better to a variety of indoor and ``Hard'' outdoor environments, which can be over 100m long, significantly outperforming the single-domain policies (Fig.~\ref{fig:qualitative_comparison}).

\begin{table}[]
    \centering
    \begin{tabular}{lcccc}
        \toprule
        Dataset(s) & LoCoBot & Tello & Vizbot & Jackal \\ \midrule
        GS & 0.26 & 0.21 & 0.51 & 0.31 \\
        RECON & 0.62 & 0.79 & 0.26 & 0.68 \\
        \ours{Ours} & \textbf{0.96} & \textbf{0.99} & \textbf{0.93} & \textbf{0.94} \\ \bottomrule
    \end{tabular}
    \caption{\textbf{Summary of navigation across robots.} A single policy trained on \MethodName-Mid outperforms the best single-robot policy for \emph{each} robot used in our experiments, mean success rate reported.}
    \label{tab:results_summary}
    \vspace*{-1.6em}
\end{table}

\begin{table}
\centering
{\footnotesize
\begin{tabular}{lcccc}
\toprule
Dataset(s) & \# & \multicolumn{2}{c}{Indoor} & Outdoor \\
\cmidrule(lr){3-4} & & Easy & Moderate & \\ \midrule
CoryHall & 1 & 0.22 & 0.13 & 0.29 \\
GS & 1 & 0.25 & 0.16 & 0.44 \\
--"-- +ImageNet & 1 & 0.35 & 0.35 & 0.57 \\
\ours\MethodName-Small & 2 & 0.82 & 0.59 & \textbf{1.0} \\
\ours\MethodName-Mid & 4 & \textbf{1.0} & \textbf{0.97} & 0.83 \\
\ours\MethodName-Large & 6 & \textbf{1.0} & \textbf{1.0} & 0.83 \\ %
\bottomrule
\end{tabular}}
\caption{\textbf{Navigation success rates on a LoCoBot.} \MethodName omnipolicies (green) result in increasingly capable navigation, in both indoor and outdoor enviroments, on an \emph{unseen} robot.}
\label{tab:results_locobot}
\vspace*{-0.8em}
\end{table}

\begin{table} %
\centering
{\footnotesize
\begin{tabular}{lcccc} %
\toprule
Dataset(s) & \# & \multicolumn{2}{c}{Outdoor} & Indoor \\
\cmidrule(lr){3-4} & & Easy & Hard & \\ \midrule
GS & 1 & 0.25 & 0.05 & 0.40 \\
RECON & 1 & 0.67 & 0.48 & 0.36 \\
--"-- +ImageNet & 1 & 0.72 & 0.52 & 0.31 \\
\ours\MethodName-Small & 2 & 0.75 & 0.52 & 0.42 \\
\ours\MethodName-Mid & 4 & \textbf{1.0} & \textbf{1.0 }& \textbf{0.82} \\
\ours\MethodName-Large & 6 & \textbf{1.0} & \textbf{1.0} & \textbf{0.88} \\
\bottomrule
\end{tabular}}
\caption{\textbf{Navigation success rates on a Jackal.} By leveraging heterogeneous datasets, \MethodName omnipolicies (green) can drive a Jackal \emph{better} than a policy trained on a Jackal-specific dataset (RECON),
also generalizing to novel indoor environments.}
\label{tab:results_jackal}
\vspace*{-1.3em}
\end{table}

\begin{figure*}
\centering
{\footnotesize
\begin{tabular}{lccc}
\toprule
Action Space & Easy & Moderate \\ \midrule
Velocities & 0.73 & 0.54 \\
Waypoints & 0.42 & 0.26 \\
\ours Norm. Waypt.   & \textbf{1.0} & \textbf{0.95} \\
\bottomrule
\end{tabular}}\;\;
{\footnotesize
\begin{tabular}{lccc}
\toprule
Architecture & Easy & Moderate \\ \midrule
Stacked~\cite{chaplot2020nts} & 0.52 & 0.72 \\
Siamese~\cite{hirose2019deep, shah2020ving, yadav2022ovrl} & 0.73 & 0.26 \\
\ours{Conditioned}~\cite{shah2021rapid, nair19ccrig} & \textbf{1.0} & \textbf{0.95} \\
\bottomrule
\end{tabular}}\;\;
{\footnotesize
\begin{tabular}{lccc}
\toprule
Context & Easy & Moderate & Hard \\ \midrule
None & \textbf{1.0} & 0.79 & 0.36 \\
\ours{Static} & \textbf{1.0} & 0.86 & 0.5 \\
\ours{Temporal}  & \textbf{1.0} & \textbf{0.92} & \textbf{0.7} \\
\bottomrule
\end{tabular}}
\captionof{table}{A systematic analysis of the design choices in Sec.~\ref{sec:analysis} reveals that choosing the right action representation (left), goal-conditioned architecture (center), and conditioning on embodiment context (right) are really important to facilitate multi-robot learning.}
\label{tab:analysis}
\vspace*{-1.2em}
\end{figure*}

\subsection{A Systematic Analysis of the Design Space}
\label{sec:analysis}

Towards answering \textbf{Q3}, we perform a systematic analysis of the design choices presented in Sec.~\ref{sec:method_description}. We evaluate each design choice on a LoCoBot, which is an \emph{unseen robot} with no corresponding training data, in indoor environments with varying levels of complexity, where ``Easy'' environments have wide passages and smooth turns, ``Moderate'' environments have tight passages or sharp turns, and ``Hard'' environments are larger (up to 50m) with a combination of tight passages and multiple turns.

\subsubsection{Shared Action Space}
\label{sec:analysis_abstraction}

We compare the three action spaces discussed in Sec.~\ref{sec:abstraction} by training three different policies on \MethodName-Mid and evaluating them in 10 environments (Table~\ref{tab:analysis}). While using velocities as an action space works well for most easy environments, often outperforming the policy using waypoints, %
both these policies struggle in environments requiring dynamic maneuvers like sharp turns. A policy based on normalized waypoints, on the other hand, significantly outperforms the others, including in the challenging environments. This suggests that normalizing the action space indeed allows the policies to learn more effectively and generalize to new robots.

\subsubsection{Embodiment Context} 
\label{sec:analysis_embodiment}

We consider two ways to represent the embodiment context: (i) temporally consistent context containing $k$ consecutive past observations $\{o_{(t-k):(t-1)}\}$, and (ii) static context,
containing a \emph{fixed} set of $k$ past observations from the robot in the target environment.
Comparing these choices in environments of varying complexities (Table~\ref{tab:analysis}), we find that adding either form of context significantly boosts the navigation performance in the harder environments, which require the robot to navigate tight passages with multiple obstacles and sharp turns. This suggests that the context helps the polices generalize better due to the additional information about the embodiment (e.g., viewpoint, speed etc.). Between the two, we found the temporal variant superior, suggesting that the temporal information (e.g., speed, turning radius etc.) is important to enable this generalization. In our main experiments discussed in Sec.~\ref{sec:results_main} and Fig.~\ref{fig:qualitative_all}, we use a temporally consistent context with $k=5$.

\subsubsection{Policy Architecture}
\label{sec:architecture}

We also compared different policy architectures to encode the goal information: (i) single-encoder stacking, where the observation and goal images are stacked along the channel dimension~\cite{chaplot2020nts}, (ii) a Siamese architecture, where the images are processed with independent encoders and the resulting embeddings are combined~\cite{hirose2019deep, shah2020ving, yadav2022ovrl}, and (iii) the conditional architecture in Fig.~\ref{fig:model}, with an additional pathway from the observation to the policy outputs~\cite{shah2021rapid, nair19ccrig}. We found that the choice of architecture significantly affects the navigation performance, with the conditional model being the most performant. We hypothesize that this is due to the additional pathway that allows the learned embeddings to be conditioned on the current observations, leading to more generalizable representations, as studied in prior work~\cite{shah2021rapid}. %

\subsection{Robustness to Degradation}
\label{sec:degradation}

\begin{figure}
    \centering
    \includegraphics[width=0.85\columnwidth]{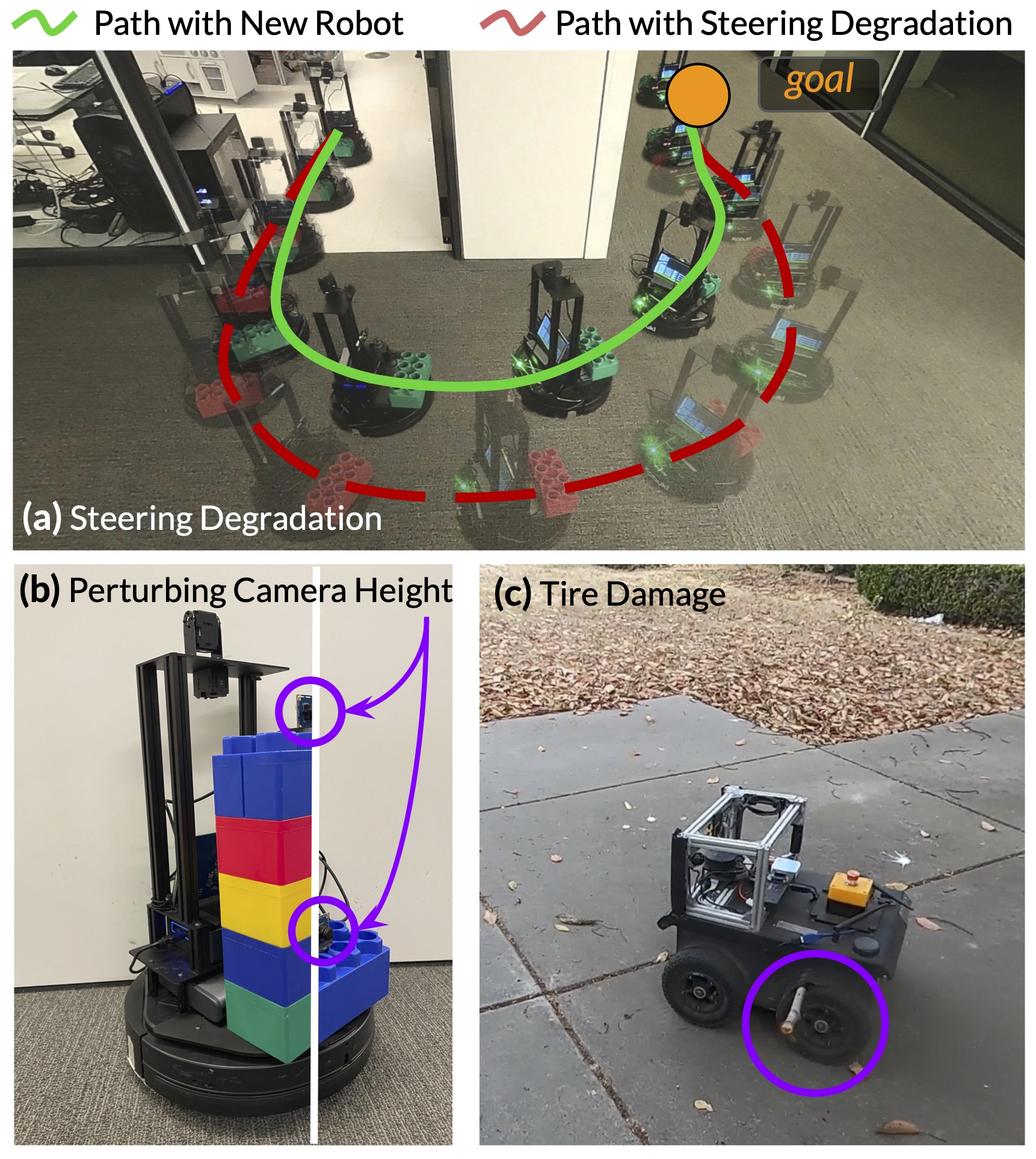} \\ \vspace{3mm}
    {\footnotesize
    \begin{tabular}{lccc}
    \toprule
        (d) & Steering & Viewpoint & Physical \\ \midrule
        Single-Domain Policy & 0.30 & 0.17 & 0.81 \\
        \ours \MethodName Policy (Ours) & \textbf{0.89} & \textbf{0.81} & \textbf{1.0} \\ \bottomrule
    \end{tabular}}
    \caption{Policies trained with \MethodName are more robust to degradation in parameters such as (a) actuation, (b) perturbed sensor viewpoint, and (c) physical damage, than single-domain policies (d).}
    \label{fig:degradation}
    \vspace*{-1.2em}
\end{figure}

A key strength of training on heterogeneous datasets is that learning across varied parameters encourages the policy to learn shared affordances across robots, thus being robust to small variation in robot parameters, such as sensor placement and mechanical properties. We show that the shared \MethodName can indeed offer such robustness by testing it under some example degradation scenarios shown in Fig.~\ref{fig:degradation}.

When testing the trained policy with a steering degradation (Fig.~\ref{fig:degradation}a), where the robot's maximum angular velocity is clipped, we find that the \MethodName can compensate for the degradation by taking a longer, smoother path towards the goal without any localization failures.
We also tested the \MethodName while perturbing the position of the camera and physically affecting the dynamics by damaging the robot \emph{during navigation}, %
and find that it can successfully reach the goals despite the degradation (Fig.~\ref{fig:degradation}d). Please see the supplemental video for these experiments.

\section{Discussion}

In this paper, we demonstrated that a general goal-conditioned navigation policy trained from navigation datasets collected by multiple distinct robots, ranging from RC cars to ATVs, can control \emph{new} robots in challenging environments. The design of our learning framework is simple, and largely follows prior work: the novel observation is that a set of relatively simple decisions, such as including a temporal context and standardizing the action space, is sufficient to enable broad generalization from heterogeneous data. Empirically, we show that our approach can enable real-world navigation for a range of robots, including some not seen in training, and even an underactuated quadrotor.

Our specific instantiation of this principle does have some limitations. Most prominently, our system does not explicitly account for differences in capabilities: we assume all robots are ground robots (though we study generalization to a quadrotor) with a forward-facing RGB camera. Handling diverse sensing, actuation (beyond variability in speed and steering), and traversability, would be an exciting direction for future work. Secondly, our dataset could be much larger: while we observe exciting generalization from 60 hours of data, a much larger and broader dataset could enable even better generalization in the future.

The promise of such a \emph{general} navigation model trained on diverse data is that it may provide a pre-trained \emph{base} model for a variety of downstream navigation applications. In the same way that computer vision researchers and practitioners typically start off by downloading a pre-trained backbone to use for their task, we hope that future navigation projects might use a pre-trained navigational omnipolicy that generalizes broadly enough to offer a ``universal'' starting point.

\section*{Acknowledgments}
This research was supported by the DARPA RACER program, ARO W911NF-21-1-0097, ARL DCIST CRA W911NF-17-2-0181, AFOSR FA9550-22-1-0273, Toyota Motor North America, and Toyota Research Institute. The authors would like to thank Haresh Karnan, Xuesu Xiao, Gregory Kahn, Xiangyun Meng, and Byron Boots, for their help in aggregating the heterogeneous dataset used for training the \MethodName. The authors would also like to thank Brian Ichter, Antonio Loquercio, Jie Tan, Devendra Singh Chaplot, Tingnan Zhang, Laura Smith, Nick Rhinehart, Frederik Ebert, and Kelvin Xu, for useful discussions and feedback on an earlier draft of the paper.

\hypersetup{linkcolor=Red, urlcolor=Blue}
\balance
\bibliographystyle{IEEEtran}
\bibliography{references}

\end{document}